\documentclass[letterpaper, 10 pt, conference]{ieeeconf}  

\IEEEoverridecommandlockouts                              

\usepackage{graphics} 
\usepackage{graphicx}
\usepackage{amsmath} 
\usepackage{amssymb}  
\usepackage{amsfonts}
\usepackage{lipsum}
\usepackage{graphicx}
\usepackage{makecell}
\usepackage{multirow}
\usepackage{tabularx}
\usepackage{float}
\usepackage[table,dvipsnames]{xcolor}
\usepackage{booktabs}
\usepackage{hyperref}

\usepackage{censor}

\usepackage{cuted}
\usepackage{capt-of}
\usepackage{afterpage}

\renewcommand{\arraystretch}{0.85} 
\title{\LARGE \bf
Robust Tightly-Coupled Filter-Based Monocular Visual-Inertial State Estimation and Graph-Based Evaluation for Autonomous Drone Racing
}

\author{Maulana Bisyir Azhari, Donghun Han, Sung Jun Park, and David Hyunchul Shim
\thanks{This paper has been accepted for publication at the 2026 IEEE/RSJ International Conference on Intelligent Robots and Systems (IROS 2026).}
\thanks{This research was financially supported by the Institute of Civil Military Technology Cooperation funded by the Defense Acquisition Program Administration and Ministry of Trade, Industry and Energy of Korean government under grant No. UM22206RD3.}
\thanks{All authors are with the Unmanned Systems Research Group (USRG), School of Electrical Engineering, Korea Advanced Institute of Science and Technology (KAIST), Daejeon 34141, South Korea. {\tt\small \{mbazhari, donghun.han, sj2park, hcshim\}@kaist.ac.kr}}
}

\begin{document}

\maketitle
\thispagestyle{empty}
\pagestyle{empty}

\begin{abstract}
Autonomous drone racing (ADR) demands state estimation that is simultaneously computationally efficient and resilient to the perceptual degradation experienced during extreme velocity and maneuvers. 
Traditional frameworks typically rely on conventional visual-inertial pipelines with loosely-coupled gate-based Perspective-n-Points (PnP) corrections that suffer from a rigid requirement for four visible features and information loss in intermediate steps. 
Furthermore, the absence of GNSS and Motion Capture systems in uninstrumented, competitive racing environments makes the objective evaluation of such systems remarkably difficult. 
To address these limitations, we propose ADR-VINS, a robust, monocular visual-inertial state estimation framework based on an Error-State Kalman Filter (ESKF) tailored for autonomous drone racing. 
Our approach integrates direct pixel reprojection errors from gate corners features as innovation terms within the filter. 
By bypassing intermediate PnP solvers, ADR-VINS maintains valid state updates with as few as two visible corners and utilizes robust reweighting instead of RANSAC-based schemes to handle outliers, enhancing computational efficiency. 
Furthermore, we introduce ADR-FGO, an offline Factor-Graph Optimization framework to generate high-fidelity reference trajectories that facilitate post-flight performance evaluation and analysis on uninstrumented, GNSS-denied environments.
The proposed system is validated using TII-RATM dataset, where ADR-VINS achieves an average RMS translation error of 0.143 m, while ADR-FGO yields 0.060 m as a smoothing-based reference.
Finally, ADR-VINS was successfully deployed in the A2RL Drone Championship Season 2, maintaining stable and robust estimation despite noisy detections during high-agility flight at top speeds of 20.9 m/s.
We further utilize ADR-FGO for post-flight performance evaluation in high-stakes, uninstrumented racing environments.

Supplementary video: \href{https://youtu.be/nzC4S16EFcQ}{https://youtu.be/nzC4S16EFcQ}
\end{abstract}

\section{Introduction}
Autonomous drone racing (ADR) has emerged as a premier benchmark for high-speed robotics, pushing the limits of perception, state estimation, and control. 
Beyond its competitive appeal, ADR serves as a proxy for time-critical real-world applications such as search and rescue, rapid medical delivery, and high-speed interception~\cite{hanover2024adr_survey}. 
In ADR, micro-aerial vehicles (MAVs) must navigate complex gate sequences at velocities exceeding 20 m/s, where latencies or drifts can result in catastrophic failure.
\begin{figure}[t!]
\vspace{8pt}
\begin{center}
\includegraphics[width=1.0\columnwidth]{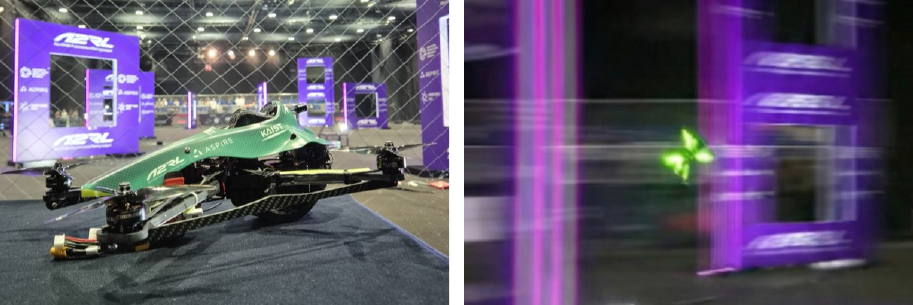}
\vspace{-18pt}
\caption{Left: The autonomous drone racing platform used in the A2RL Drone Championship Season 2, featuring a low-cost monocular RGB camera and IMU sensors. Right: Autonomous racing flight at high speeds over 20 m/s featuring the proposed ADR-VINS for its real-time state estimation.}
\label{fig:fig1}
\end{center}
\vspace{-26pt}
\end{figure}
The primary technical challenge lies in maintaining robust state estimation amidst extreme maneuvers—which induce motion blur, light flickers, and significant vibrations—while operating under limited computational and power constraints of small platforms, such as shown in Fig. \ref{fig:fig1}, which was used in A2RL Drone Championship Season 2\footnote[1]{https://a2rl.io/autonomous-drone-race}.

\begingroup
\renewcommand{\arraystretch}{0.8} 
\setlength{\aboverulesep}{1pt}
\setlength{\belowrulesep}{1pt}

\begin{table*}[ht] 
    \centering 
    \caption{Comparison of Gate-Based State-Estimation Methods for Autonomous Drone Racing.}
    \scriptsize
    \vspace{-10pt}
    \label{tab:gate_based_methods} 
    \setlength{\tabcolsep}{4pt}
    \begin{tabular}{cccccccccccccccc} 
        \midrule \midrule
        \multirow{2}{*}{\textbf{Ref}} &{}
        &\multicolumn{2}{c}{\textbf{Validation}} &{}
        &\multicolumn{3}{c}{\textbf{Sensors}} &{}
        &\multicolumn{3}{c}{\textbf{Gate/Corner Detection}} &{}
        &\multirow{2}{*}{\textbf{State Estimation Pipeline}} &{}
        &\multirow{2}{*}{\textbf{Top Speed}} \\
        &{} &Real-World &Dataset &{} &Camera &IMU &Additional &{} &Gate &Corner &Type &{}\\
        \cmidrule{1-1} \cmidrule{3-4} \cmidrule{6-8} \cmidrule{10-12} \cmidrule{14-14} \cmidrule{16-16}
        \cite{foehn2022alphapilot_uzh} &{} &AIRR$^*$ &- &{} &Stereo &Yes &1D LiDAR &{} &- &U-Net &1-stage$^\dagger$ &{} &ROVIO + Drift EKF &{} &8 m/s \\
        \cite{de2022mavlab_airr} &{} &AIRR$^*$ &- &{} &Mono &Yes &1D LiDAR &{} &GateNet &Snake-Gate &2-stages &{} &RANSAC + PnP + MHE &{} &9.2 m/s \\
        \cite{kaufmann2023champion} &{} &Self-hosted &- &{} &Stereo &Yes &- &{} &- &U-Net &1-stage$^\dagger$ &{} &T265-VIO + PnP + Drift KF &{} &22 m/s \\
        \cite{bosello2025on_your_own} &{} &Self-hosted &- &{} &Stereo &Yes &- &{} &YoloV8 &U-Net &2-stages &{} &T265-VIO + PnP + Drift KF + EKF &{} &21.1 m/s\\
        \cite{novák2026visiononlyuavstateestimation} &{} &A2RL S1$^*$ &- &{} &Mono &Yes &- &{} &- &U-Net &1-stage$^\dagger$ &{} &VINS-Mono/OpenVINS + PnP + Drift KF &{} &12.5 m/s \\
        \cite{azhari2025driftcorrected_adr} &{} &A2RL S1$^*$ &- &{} &Mono &Yes &- &{} &\multicolumn{2}{c}{YoloV8-Pose} &1-stage$^\ddagger$ &{} &OpenVINS + PnP + Drift KF &{} &12.2 m/s\\
        \cite{bahnam2026monorace} &{} &A2RL S1$^*$ &- &{} &Mono &Yes &Motor RPM &{} &GateNet &LSD &2-stages &{} &RANSAC + PnP + EKF &{} &28.2 m/s\\
        \midrule
        Ours &{} &A2RL S2$^*$ &TII-RATM &{} &Mono &Yes &- &{} &\multicolumn{2}{c}{RTMO} &1-stage$^\ddagger$ &{} &ESKF &{} &20.9 m/s\\
        \midrule \midrule
    \end{tabular}
    \vspace{-4pt}
    \begin{itemize}
        \item[$^*$] Competition-based real-world validation. AIRR: Alphapilot's Artificial Intelligence Robotic Racing 2019, A2RL S1: A2RL Drone Championship Season 1 (April 2025), A2RL S2: A2RL Drone Championship Season 2 (January 2026). Ours provide both competition-based and open-source dataset-based validation.
        \item[$^\dagger$] Bottom-up approach which detects the corner by segmentation then associate the corners within the same gate via part-affinity fields (PAFs).
        \item[$^\ddagger$] Top-down approach which detects gate and corner simultaneously in one model.
    \end{itemize}
    \vspace{-20pt}
\end{table*}
\endgroup

The current State-of-the-Art (SOTA) navigational frameworks in ADR often adopt a cascaded or loosely-coupled architecture ~\cite{foehn2022alphapilot_uzh,de2022mavlab_airr,kaufmann2023champion,bosello2025on_your_own,novák2026visiononlyuavstateestimation,azhari2025driftcorrected_adr}. These systems typically utilize a standalone Visual-Inertial Navigation System (VINS) or Odometry (VIO) module for ego-motion estimation~\cite{bloesch2015rovio,geneva2020openvins,qin2018vinsmono}, and correct the accumulated drift using Perspective-n-Point (PnP) pose estimations relative to known gates. 
However, this approach has several limitations. 
First, it places a heavy reliance on the underlying VIO performance, which can degrade rapidly during aggressive maneuvers. 
Second, the PnP step requires a strict minimum of four visible features for a unique solution—often violated by occlusions or sharp turns—and exhibits high sensitivity to detection noise, often leading to increased latency through RANSAC-based filtering. 
Furthermore, reducing multiple 2D pixel measurements into a single 3D pose before filter integration discards significant geometric information.

Furthermore, research progress within the ADR community is significantly constrained by the high cost and logistical overhead of "gold-standard" large-scale Motion Capture (MoCap) systems. 
Consequently, while recent autonomous systems have demonstrated human-competitive performance~\cite{kaufmann2023champion} even in uninstrumented environments~\cite{bosello2025on_your_own}, their evaluation often relies on end-to-end mission success, omitting the standardized, quantitative metrics for detection, state estimation, and control precision. 
This gap leaves a question unanswered: \textit{what level of precision is truly "good enough" for superhuman racing?}
Without a verifiable reference, evaluating system performance in uninstrumented environments remains difficult.

To address these challenges, the primary contributions of this work are as follows:
\begin{itemize}
    \item \textbf{ADR-VINS}: A tightly-coupled monocular visual-inertial state estimator based on an Error-State Kalman Filter (ESKF)~\cite{sola2017quaternion_eskf} that directly integrates gate corner reprojection errors as innovation terms. By eliminating intermediate PnP solvers and RANSAC, the method enables valid updates with as few as two visible corners under high-speed racing conditions.
    \item \textbf{ADR-FGO}: An offline factor-graph optimization framework~\cite{dellaert2017factor_graph} that performs global batch optimization to produce a trajectory-consistent reference estimate for post-flight evaluation in uninstrumented racing environments.
    \item \textbf{Validation and Deployment}: Experimental validation on the TII-RATM dataset and successful deployment in the A2RL Drone Championship Season 2, demonstrating stable estimation at speeds exceeding $20$ m/s.
\end{itemize}

\begin{figure*}[t!]
\begin{center}
\includegraphics[width=1\textwidth]{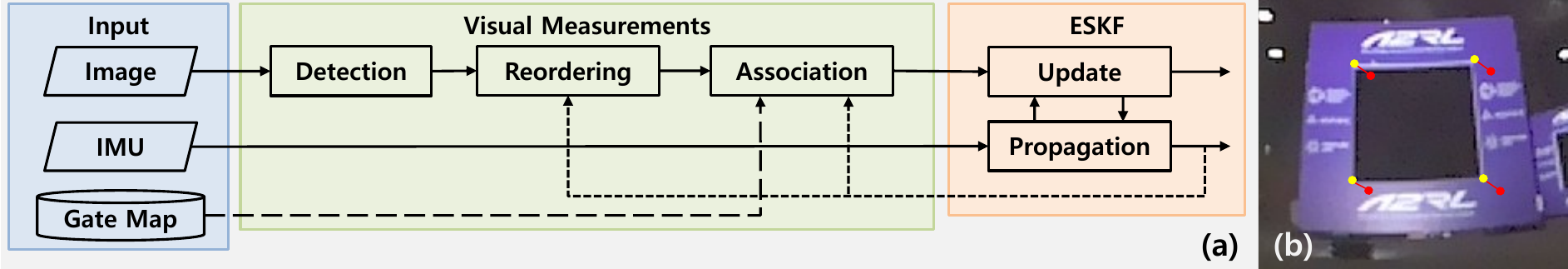}
\vspace{-22pt}
\caption{Overview of ADR-VINS. (a) The framework consists of a visual measurement pipeline (including detection, reordering, and association) and a tightly-coupled Error-State Kalman Filter (ESKF) that fuses raw IMU data with direct pixel reprojection errors to provide robust, high-rate state estimation. (b) Illustration of the corner detections (yellow dots), reprojections (red dots), and the corresponding reprojection errors (red lines).}
\vspace{-22pt}
\label{fig:adr_vins_overview}
\end{center}
\end{figure*}

\section{Related Work}
Autonomous drone racing (ADR) architectures are generally categorized into modular and end-to-end approaches~\cite{hanover2024adr_survey}. 
Modular pipelines, such as \textit{Swift}~\cite{kaufmann2023champion}, \textit{On Your Own}~\cite{bosello2025on_your_own}, and \textit{MonoRace}~\cite{bahnam2026monorace}, decouple perception from control (e.g., MPC or RL), enabling better transferability of each module across different high-level tasks.

State estimation implementations vary significantly across these systems, as summarized in Table \ref{tab:gate_based_methods}. 
Early competitions allowed for more diverse sensor suites, including stereo cameras and 1D LiDARs~\cite{foehn2022alphapilot_uzh,de2022mavlab_airr}. 
Consequently, benchmarks like \textit{Swift}~\cite{kaufmann2023champion} and \textit{On Your Own}~\cite{bosello2025on_your_own} relied on proprietary stereo vision (Intel T265), which limits transferability to different camera setups. 
Recent competitions like the A2RL Drone Championships emphasize restrictive monocular RGB and low-quality IMU setups. 
In A2RL S1, the 3rd~\cite{azhari2025driftcorrected_adr} and 4th~\cite{novák2026visiononlyuavstateestimation} place entries utilized general VIO~\cite{geneva2020openvins,qin2018vinsmono} with PnP-pose corrections, which often degrade under severe high-speed conditions. 
Conversely, the winner \textit{MonoRace}~\cite{bahnam2026monorace} employed an IMU-Pose EKF, using sub-pixel corner detection and RANSAC-based rejection prior to gate-based PnP estimation to correct IMU propagation.

Despite these advancements, PnP-based estimation remains fundamentally limited by a rigid four-feature requirement and high sensitivity to detection noise. 
Furthermore, compressing 2D pixel measurements into a single 3D pose discards underlying geometric uncertainty, while RANSAC-based rejection introduces latency and is prone to noise and initial guess. 
Although \cite{foehn2022alphapilot_uzh} utilized reprojection residuals to mitigate drift, the system remained secondary to the underlying VIO module. 
In contrast, our proposed ADR-VINS tightly couples these residuals within an ESKF, enabling robust corrections with as few as two visible corners and employing Huber-based reweighting to handle outliers without RANSAC-induced latency.

Finally, current ADR evaluations typically focus on lap times rather than granular estimation or control precision~\cite{kaufmann2023champion,bosello2025on_your_own}. 
Existing benchmarking efforts often suffer from practical limitations: Team Fly4Future~\cite{novák2026visiononlyuavstateestimation} used GNSS for outdoor evaluation, but this resulted in a reference obtained in an environment entirely distinct from the uninstrumented, GNSS-denied arenas used in competition; Team KAIST~\cite{azhari2025driftcorrected_adr} used MAPLAB~\cite{cramariuc2022maplab} for full-batch optimization, yet it requires loop-closures that are difficult to satisfy in feature-sparse high-speed racing; and Team MAVLab~\cite{bahnam2026monorace} used agreement checks between EKF and PnP estimates, but these are self-referential and may mask common-mode errors. 
To address these limitations, we propose ADR-FGO, an offline, non-causal factor-graph optimization framework~\cite{dellaert2017factor_graph}. 
By leveraging global measurements and non-causal smoothing, ADR-FGO generates a significantly more accurate trajectory, providing the reference needed to evaluate system performance in uninstrumented racing environments.

\section{ADR-VINS}
ADR-VINS is a computationally efficient, real-time monocular visual-inertial estimator tailored for autonomous drone racing by tightly coupling gate corner detections within the filter backend.
As illustrated in Fig. \ref{fig:adr_vins_overview}, the visual measurement module detects the gate's bounding box and corners simultaneously, which are then reordered to resolve switched corners and associated with the known gate map.
These visual measurements are then fused with IMU measurements in an ESKF estimator.

\subsection{Detection} 
\label{sec:gate_corner_detection}
For an incoming image $\mathbf{I}_k \in \mathbb{R}^{H \times W}$ at discrete time $k$, the model predicts a set of detections $\hat{G}_k = \{\hat{G}_{k,0}, \dots, \hat{G}_{k,N-1}\}$. 
Each detection contains the pixel coordinates $\mathbf{u}_{c \in \{TL, TR, BR, BL\}} = [u, v]^\top \in [0, W] \times [0, H]$, undistorted and normalized as $\tilde{\mathbf{u}}_{c} = [\tilde{u}, \tilde{v}]^\top$, and confidence scores $s_{c} \in [0, 1]$ for the four inner corners: top-left (TL), top-right (TR), bottom-right (BR), and bottom-left (BL).

\subsection{Reordering}
\label{sec:corner_reordering}
To resolve corner label switching due to visual ambiguities at high maneuvers (roll $> 60^\circ$) and limited training data in such scenarios, detected corners are reordered based on the current nominal state $(\mathbf{p}_{wb}, \mathbf{q}_{wb})$ and camera extrinsics $(\mathbf{R}_{bc}, \mathbf{p}_{bc})$. 
We project a gravity-aligned "up" vector $\boldsymbol{\mu}_{up}$ and "right" vector $\tilde{\boldsymbol{\mu}}_{right} = [-\tilde{\mu}_{up,v}, \tilde{\mu}_{up,u}]^\top$ into the image plane using a probe point $\mathbf{p}_{ref}^\mathcal{W}$. 
This point is obtained by back-projecting the detection centroid $\bar{\mathbf{u}}$ to a heuristic distance $d=3$ m and transforming it to the world frame:
\begin{equation}
    \begin{aligned}
    \mathbf{p}_{ref}^\mathcal{W} &= \mathbf{R}(\mathbf{q}_{wb}) (\mathbf{R}_{bc} [ \tilde{u}d, \tilde{v}d, d ]^\top + \mathbf{p}_{bc}) + \mathbf{p}_{wb} \\
    \Delta \tilde{\boldsymbol{\mu}} &= \pi(\mathbf{R}_{bc}^\top (\mathbf{R}(\mathbf{q}_{wb})^\top (\mathbf{p}_{ref}^\mathcal{W} + \mathbf{g}_z - \mathbf{p}_{wb})) - \mathbf{p}_{bc}) \\
    &- \pi(\mathbf{R}_{bc}^\top (\mathbf{R}(\mathbf{q}_{wb})^\top (\mathbf{p}_{ref}^\mathcal{W} - \mathbf{g}_z - \mathbf{p}_{wb})) - \mathbf{p}_{bc}) \\
    \tilde{\boldsymbol{\mu}}_{up} &= \Delta \tilde{\boldsymbol{\mu}}/\|\Delta \tilde{\boldsymbol{\mu}}\|,
    \end{aligned}
\end{equation}
where $\mathbf{g}_z = [0, 0, 1]^\top$ and $\pi(\cdot)$ is perspective projection to the image plane. 
Corners are then assigned to physical labels $j \in \{TL, TR, BR, BL\}$ by maximizing:
\begin{equation}
  S(j,c) = \sigma_{j}^{up}(\delta_{c} \cdot \boldsymbol{\mu}_{up}) + \sigma_{j}^{right}(\delta_{c} \cdot \boldsymbol{\mu}_{right}),  
\end{equation}
where $\delta_c = \tilde{\mathbf{u}}_c - \bar{\mathbf{u}}$ and $\sigma \in \{-1, 1\}$ denotes the target quadrant for each corner (e.g., TL: $\{+1, -1\}$).

\subsection{Association} 
\label{sec:gate_association}
Detections $\hat{G}_{k,n}$ are associated with map gates $G_{k,m} \in \mathcal{G}$ that are within the camera field of view. We solve for the optimal association $m^*$ by minimizing a cost function $J(m)$ that fuses spatial and scale consistency:
\begin{equation}
  m^* = \arg \min_{m} \frac{d_m}{\rho_m}, \quad \text{t.s. } d_{m^*} < 75\enspace \text{px} \wedge \rho_{m^*} > 0.2  
\end{equation}
where $d_m = \|C(\hat{G}_{k,n}) - C(G_{k,m})\|_2$ represents the Euclidean distance in the image plane between the detection centroid and the projected map gate centroid.
The function $C(\cdot)$ computes the center of the gates's corner coordinates. 
The $d_m$ threshold was chosen empirically to balance high-speed association validity against false positives.
The term $\rho_m$ is the area consistency ratio:
\begin{equation}
  \rho_m = \min(A(\hat{G}_{k,n})/A(G_{k,m}), \enspace A(G_{k,m})/A(\hat{G}_{k,n}))  
\end{equation}
where the function $A(\cdot)$ computes the area of the gate based on the detected corners. 
The ratio $\rho_m$ serves as a proxy for proximity, distinguishing between overlapping gates at varying depths to ensure correct association.

Once gates are associated, we resolve potential left-right flips that occur when a gate is viewed from the rear by comparing the total reprojection error.
We further discard far away gates with more than $15$ m distance from camera.
The resulting measurements at time $k$ are defined as the set $\mathcal{Z}_k = \{\mathbf{z}_{k,0}, \dots, \mathbf{z}_{k,M-1}\}$, where each measurement $\mathbf{z}_{k,i} = \{ \tilde{\mathbf{u}}_{i}, \mathbf{p}_{G,i}^\mathcal{W} \}$ represents the pair of a 2D corner detection and its associated 3D gate corner in the world frame.

\subsection{Error-State Kalman Filter}
\label{sec:eskf}
The core of ADR-VINS is an Error-State Kalman Filter (ESKF)~\cite{sola2017quaternion_eskf} that maintains a high-rate nominal state while estimating small perturbations to ensure numerical stability on the $SO(3)$ manifold. 
The filter recursively propagates the state using high-frequency IMU data and performs updates by minimizing feature residuals from corner detections.

\subsubsection{State Definition}
The nominal state $\bar{\mathbf{x}}$ is a vector comprising the body position $\bar{\mathbf{p}}$, velocity $\bar{\mathbf{v}}$, attitude quaternion $\bar{\mathbf{q}}$, and IMU biases $\bar{\mathbf{b}}_a$ and $\bar{\mathbf{b}}_\omega$:
\begin{equation}
    \bar{\mathbf{x}} = [\bar{\mathbf{p}}^\top, \bar{\mathbf{v}}^\top, \bar{\mathbf{q}}^\top, \bar{\mathbf{b}}_a^\top, \bar{\mathbf{b}}_\omega^\top]^\top
\end{equation}

The true state $\mathbf{x}$ is defined as the composition ($\oplus$) of the nominal state $\bar{\mathbf{x}}$ and the error state $\delta \mathbf{x} \in \mathbb{R}^15$:
\begin{equation}
    \mathbf{x} = \bar{\mathbf{x}}\oplus \delta \mathbf{x},
\end{equation}
where error state vector $\delta \mathbf{x} \in \mathbb{R}^{15}$ is defined as $\delta \mathbf{x} = [\delta \mathbf{p}^\top, \delta \mathbf{v}^\top, \delta \boldsymbol{\theta}^\top, \delta \mathbf{b}_a^\top, \delta \mathbf{b}_\omega^\top]^\top$.
The individual components of the true state are as follows:
\begin{equation}
\begin{split}
    &\mathbf{p} = \bar{\mathbf{p}} + \delta \mathbf{p}, \quad  \mathbf{v} = \bar{\mathbf{v}} + \delta \mathbf{v}, \quad \mathbf{q} = \bar{\mathbf{q}} \otimes \exp(\delta \boldsymbol{\theta} / 2), \\
    &\mathbf{b}_a = \bar{\mathbf{b}}_a + \delta \mathbf{b}_a, \quad \mathbf{b}_\omega = \bar{\mathbf{b}}_\omega + \delta \mathbf{b}_\omega.
\end{split}
\end{equation}
where $\exp(\cdot)$ is the exponential map from $\mathfrak{so}(3)$ to $SO(3)$, and $\delta \boldsymbol{\theta}$ represents the small attitude perturbation.

\subsubsection{IMU Propagation}
We model raw IMU measurements $a_m \in \mathbb{R}^3$ and $\omega_m \in \mathbb{R}^3$ as corrupted by additive white noise $n$ and slowly varying biases $b$:
\begin{equation}
    \mathbf{a}_m = \mathbf{R}^\top(\mathbf{a} - \mathbf{g}) + \mathbf{b}_a + \mathbf{n}_a, \quad \boldsymbol{\omega}_m = \boldsymbol{\omega} + \mathbf{b}_\omega + \mathbf{n}_\omega,
\end{equation}
where $\mathbf{n}_a \sim \mathcal{N}(0, \sigma_a^2)$ and $\mathbf{n}_\omega \sim \mathcal{N}(0, \sigma_\omega^2)$. The biases $\mathbf{b}_a$ and $\mathbf{b}_\omega$ are modeled as random walk processes driven by $\mathbf{n}_b \sim \mathcal{N}(0, \sigma_b^2)$ such that $\dot{\mathbf{b}}_a = \mathbf{n}_{b_a}$ and $\dot{\mathbf{b}}_\omega = \mathbf{n}_{b_\omega}$.

At each IMU step $\Delta t$, the nominal state is propagated using kinematic integration:
\begin{equation}
\begin{split}
    \bar{\mathbf{p}}_{k+1} &= \bar{\mathbf{p}}_k + \bar{\mathbf{v}}_k \Delta t + \frac{1}{2}(\mathbf{R}(\bar{\mathbf{q}}_{k})(\mathbf{a}_{m,k} - \bar{\mathbf{b}}_{a,k}) + \mathbf{g}) \Delta t^2 \\
    \bar{\mathbf{v}}_{k+1} &= \bar{\mathbf{v}}_k + (\mathbf{R}(\bar{\mathbf{q}}_{k})(\mathbf{a}_{m,k} - \bar{\mathbf{b}}_{a,k}) + \mathbf{g}) \Delta t \\
    \bar{\mathbf{q}}_{k+1} &= \bar{\mathbf{q}}_k \otimes \exp((\boldsymbol{\omega}_{m,k} - \bar{\mathbf{b}}_{\omega,k}) \Delta t)
\end{split}
\end{equation}
The error state dynamics are governed by $\delta \dot{\mathbf{x}} = \mathbf{F}_c \delta \mathbf{x} + \mathbf{G}_c \mathbf{w}$, where $\mathbf{G}_c$ and $\mathbf{w}$ is the noise Jacobian and vector, respectively.
The system Jacobian $\mathbf{F}_c$ is defined as:
{
\setlength{\arraycolsep}{2pt}
\begin{equation}
\mathbf{F}_c = \begin{bmatrix} 
    \mathbf{0} & \mathbf{I} & \mathbf{0} & \mathbf{0} & \mathbf{0} \\[-2pt]
    \mathbf{0} & \mathbf{0} & -[\mathbf{R}(\bar{\mathbf{q}})(\mathbf{a}_m - \bar{\mathbf{b}}_a)]_\times & -\mathbf{R}(\bar{\mathbf{q}}) & \mathbf{0} \\[-2pt]
    \mathbf{0} & \mathbf{0} & -[\boldsymbol{\omega}_m - \bar{\mathbf{b}}_\omega]_\times & \mathbf{0} & -\mathbf{I} \\[-2pt]
    \mathbf{0} & \mathbf{0} & \mathbf{0} & \mathbf{0} & \mathbf{0} \\[-2pt]
    \mathbf{0} & \mathbf{0} & \mathbf{0} & \mathbf{0} & \mathbf{0} 
    \end{bmatrix}
\end{equation}
}
where $\mathbf{I}\in\mathbb{R}^{3\times3}$ and $\mathbf{0}\in\mathbb{R}^{3\times3}$ are identity and zero matrices, respectively.
For discrete-time implementation, we compute the transition matrix $\mathbf{F}_k \approx \mathbf{I} + \mathbf{F}_c \Delta t$. The discrete process noise matrix $\mathbf{Q}_k \in \mathbb{R}^{15 \times 15}$ is constructed as a block-diagonal matrix:
\begin{equation}
\mathbf{Q}_k = \begin{bmatrix}
    \mathbf{Q}_p & \mathbf{Q}_{pv} & \mathbf{0} & \mathbf{0} & \mathbf{0} \\[-1pt]
    \mathbf{Q}_{pv}^\top & \mathbf{Q}_v & \mathbf{0} & \mathbf{0} & \mathbf{0} \\[-1pt]
    \mathbf{0} & \mathbf{0} & \mathbf{Q}_\theta & \mathbf{0} & \mathbf{0} \\[-1pt]
    \mathbf{0} & \mathbf{0} & \mathbf{0} & \mathbf{Q}_{b_a} & \mathbf{0} \\[-1pt]
    \mathbf{0} & \mathbf{0} & \mathbf{0} & \mathbf{0} & \mathbf{Q}_{b_\omega}
    \end{bmatrix}
\end{equation}
where the individual noise blocks are: $\mathbf{Q}_p = \mathbf{I} \cdot\sigma_a^2 \Delta t^3/3$, $\mathbf{Q}_{pv} = \mathbf{I} \cdot \sigma_a^2 \Delta t^2/2$, $\mathbf{Q}_v = \mathbf{I}\cdot\sigma_a^2 \Delta t$, $\mathbf{Q}_\theta = \mathbf{I} \cdot \sigma_\omega^2 \Delta t$, $\mathbf{Q}_{b_a} = \mathbf{I} \cdot \sigma_{b_a}^2 \Delta t$, and $\mathbf{Q}_{b_\omega} = \mathbf{I} \cdot \sigma_{b_\omega}^2 \Delta t$. 
The error covariance $\mathbf{P}$ is then propagated as:
\begin{equation}
    \mathbf{P}_{k+1} = \mathbf{F}_k \mathbf{P}_k \mathbf{F}_k^\top + \mathbf{Q}_k
\end{equation}
\subsubsection{Update}
For each measurement $z_{k,i} \in \mathcal{Z}_k$, the measurement residual $\mathbf{r}_{k,i}$ is defined as:
\begin{align}
    \mathbf{r}_{k,i} = \tilde{\mathbf{u}}_{k,i} - h(\bar{\mathbf{x}}_k, \mathbf{p}_{G,i}^\mathcal{W})
    \label{eq:eskf_measurements}
\end{align}
The observation model $h(\cdot)$ projects the world-frame landmark into normalized image coordinates via:
\begin{equation}
    \begin{aligned}
        h(\bar{\mathbf{x}}_k, \mathbf{p}_{G,i}^\mathcal{W}) &= \pi(\mathbf{p}^\mathcal{C}_{k,i}) \\ 
        &= \pi\left(\mathbf{R}_{bc}^\top \left(\mathbf{R}(\bar{q})^\top (\mathbf{p}_{G,i}^\mathcal{W} - \bar{\mathbf{p}}) - \mathbf{p}_{bc}\right)\right)
    \end{aligned}
\end{equation}
where $\mathbf{p}^\mathcal{C}_{k,i} = [x^\mathcal{C}, y^\mathcal{C},z^\mathcal{C}]^\top$ is the corner landmark in the camera frame. 
The measurement Jacobian $\mathbf{H}_{k,i} $ is defined as:
\begin{equation}
    \begin{aligned}
        \mathbf{H}_{k,i} &= \frac{\partial h}{\partial \delta \mathbf{x}} = \frac{\partial \pi}{\partial \mathbf{p}^\mathcal{C}_{k,i}} \cdot \frac{\partial \mathbf{p}^\mathcal{C}_{k,i}}{\partial \delta \mathbf{x}}, \quad \text{where:} \\
        \frac{\partial \pi}{\partial \mathbf{p}^\mathcal{C}_{k,i}} &= \begin{bmatrix} 1/z^\mathcal{C} & 0 & -x^\mathcal{C}/(z^\mathcal{C})^2 \\ 0 & 1/z^\mathcal{C} & -y^\mathcal{C}/(z^\mathcal{C})^2 \end{bmatrix} \\
        \frac{\partial \mathbf{p}^\mathcal{C}_{k,i}}{\partial \delta \mathbf{x}} &= [ 
-\mathbf{R}_{bc}^\top \mathbf{R}(\bar{q})^\top  \enspace \mathbf{0}_{3\times3} \enspace \cdots\\ 
&\quad \quad \mathbf{R}_{bc}^\top [\mathbf{R}(\bar{\mathbf{q}})^\top (\mathbf{p}_{G,i}^\mathcal{W} - \bar{\mathbf{p}})]_\times \enspace \mathbf{0}_{3\times3} \enspace \mathbf{0}_{3\times3} ]
    \end{aligned}
\end{equation}

To handle outliers without RANSAC latency, we implement robust reweighting using a Huber function. The innovation covariance $\mathbf{S} = \mathbf{H} \mathbf{P} \mathbf{H}^\top + \mathbf{R}_{cov}$ is used to compute the Mahalanobis distance $e = \sqrt{r^\top \mathbf{S}^{-1} r}$. 
The measurement covariance $\mathbf{R}_{cov}$ is then inflated by $w$:
\begin{equation}
    w = \operatorname{min}(1.0, \frac{\tau_e}{e}), \quad \mathbf{R}_{eff,k,i} = \frac{1}{w} \mathbf{R}_{cov,k,i}
\end{equation}

The nominal state $\bar{\mathbf{x}}$ is updated using the Kalman gain $\mathbf{K}$ and error perturbation $\delta \mathbf{x}$:
\begin{equation}
\begin{aligned}
    \mathbf{K}_{k,i} &= \mathbf{P}_k \mathbf{H}_{k,i}^{\top}(\mathbf{H}_{k,i}\mathbf{P}_k\mathbf{H}_{k,i}^{\top} + \mathbf{R}_{eff,k,i})^{-1} \\
    \bar{\mathbf{x}}_k &\leftarrow \bar{\mathbf{x}}_k \oplus \delta \mathbf{x}_k, \quad \text{where:}\enspace \delta \mathbf{x}_k = \mathbf{K}_{k,i} \mathbf{r}_k\\
     \mathbf{P}_k &\leftarrow (\mathbf{I} - \mathbf{K}_{k,i}\mathbf{H}_{k,i})\mathbf{P}_k
     \label{eskf_update}
\end{aligned}  
\end{equation}

The error state $\delta \mathbf{x}$ is reset to zero after each injection. 
The entire process of (\ref{eq:eskf_measurements})-(\ref{eskf_update}) is repeated sequentially for all available measurements $z_{k,i} \in \mathcal{Z}_k$.

\begin{figure}[t!]
\vspace{2pt}
\begin{center}
\includegraphics[width=1.0\columnwidth]{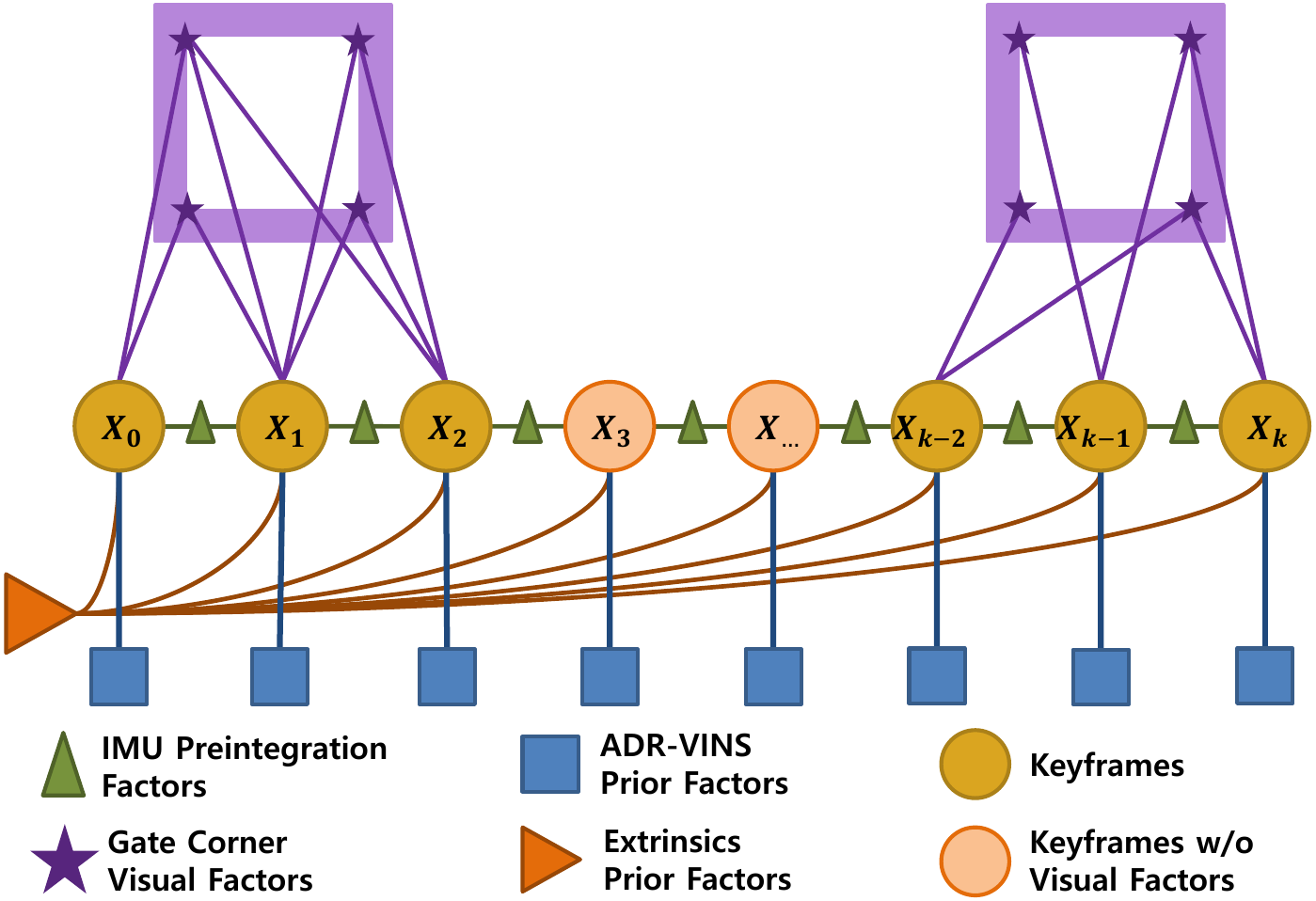}
\vspace{-18pt}
\caption{The proposed factor graph structure in ADR-FGO. The graph integrates IMU pre-integration, gate corner detections, pose prior factors from ADR-VINS, and camera extrinsics priors. Keyframes are maintained even during visual outages to distribute drift and IMU biases across these visual-less sections to provide better interpolation.}
\label{fig:adr_fgo_graph}
\end{center}
\vspace{-18pt}
\end{figure}

\begingroup

\renewcommand{\arraystretch}{0.75} 
\setlength{\aboverulesep}{1.0pt}
\setlength{\belowrulesep}{1.0pt}

\begin{table*}[t!]
    \centering
    \caption{Results on the TII-RATM Dataset.}
    \vspace{-8pt}
    \label{tab:tii_results}
    \setlength{\tabcolsep}{4.5pt}
    \begin{tabular}{cccccccccccccccccccccccccccccc}
    \midrule \midrule
    \multicolumn{2}{c}{\multirow{3}{*}{Seq}}    
            &{} &\multicolumn{15}{c}{\textbf{ONLINE METHODS}}
            &{} &\multicolumn{7}{c}{\textbf{OFFLINE METHODS}} \\
    \cmidrule{4-18} \cmidrule{20-27}
    {}  &{} &{}
        &\multicolumn{3}{c}{OpenVINS} &{}
        &\multicolumn{3}{c}{PnP-Pose} &{} 
        &\multicolumn{3}{c}{PnP+EKF} &{}
        &\multicolumn{3}{c}{\textbf{ADR-VINS}} &{}
        &\multicolumn{3}{c}{MAPLAB} &{}
        &\multicolumn{3}{c}{\textbf{ADR-FGO}} &{}\\

    & &
    &$e_t$ &$e_r$ &$e_v$ &{}
    &$e_t$ &$e_r$ &$e_v$ &{}
    &$e_t$ &$e_r$ &$e_v$ &{}
    &$e_t$ &$e_r$ &$e_v$ &{}
    &$e_t$ &$e_r$ &$e_v$ &{}
    &$e_t$ &$e_r$ &$e_v$ \\
    \midrule

    \multirow{4}{*}{\rotatebox{90}{\parbox{30pt}{\centering {Piloted}}}}
    &05P &{}
        &.296 &1.77 &.293 &{}
        &.466 &4.05 &- &{}
        &.282 &1.59 &.388 &{}
        &.130 &1.73 &.202 &{}
        
        &.194 &1.60 &.153 &{}
        &.052 &1.16 &.090 \\
        
    &06P &{}
        &.943 &2.21 &.449 &{}
        &.488 &4.17 &- &{}
        &.350 &2.64 &.505 &{}
        &.134 &1.69 &.218 &{}
        
        &.159 &1.35 &.132 &{}
        &.061 &1.47 &.100 \\
        
    &11P &{}
        &.369 &3.12 &.345 &{}
        &.398 &3.39 &- &{}
        &.242 &1.76 &.371 &{}
        &.132 &1.85 &.199 &{}

        &.249 &2.20 &.216 &{}
        &.054 &1.25 &.110 \\
    
    &12P &{}
        &.204 &2.39 &.194 &{}
        &.298 &3.87 &- &{}
        &.245 &1.62 &.355 &{}
        &.169 &2.02 &.221 &{}
        
        &.297 &1.67 &.208 &{}
        &.049 &1.39 &.090 \\

    \cmidrule{2-27}
    \multicolumn{2}{c}{Avg @P} &{}
        &.453 &2.37 &.321 &{}
        &.412 &3.80 &- &{}
        &.280 &1.90 &.405 &{}
        &.141 &1.82 &.210 &{}
        
        &.224 &1.71 &.177 &{} 
        &.054 &1.32 &.098 \\
    
    \midrule

    \multirow{6}{*}{\rotatebox{90}{\parbox{40pt}{\centering {Autonomous}}}}
    &05A &{}
        &.529 &5.86 &.754 &{}
        &.997 &5.23 &- &{}
        &.498 &3.48 &.923 &{}
        &.132 &2.10 &.329 &{}
        
        &.439 &4.58 &.674 &{}
        &.069 &1.97 &.141 \\
    
    &06A &{}
        &.288 &4.86 &.507 &{}
        &.944 &4.90 &- &{}
        &.411 &3.49 &.910 &{}
        &.132 &2.21 &.325 &{}
        
        &.544 &7.84 &.675 &{}
        &.068 &2.16 &.162 \\
    
    &11A &{}
        &.615 &4.27 &.504 &{}
        &.998 &6.71 &- &{}
        &.553 &4.12 &.751 &{}
        &.107 &2.34 &.218 &{}
        
        &.931 &9.14 &.996 &{}
        &.058 &2.33 &.139 \\
        
    &12A &{}
        &.329 &4.70 &.428 &{}
        &.907 &6.16 &- &{}
        &.534 &4.64 &.864 &{}
        &.099 &2.54 &.248 &{}
        
        &.304 &4.51 &.386 &{}
        &.061 &2.53 &.155 \\
        
    &17A &{}
        &.865 &2.57 &.592 &{}
        &1.34 &6.08 &- &{}
        &.394 &2.68 &.651 &{}
        &.191 &1.97 &.423 &{}
        
        &1.02 &4.77 &.844 &{}
        &.068 &1.81 &.195 \\
        
    &18A &{}
        &.723 &3.21 &.765 &{}
        &1.23 &6.18 &- &{}
        &.584 &2.65 &.724 &{}
        &.208 &2.17 &.449 &{}
        
        &.773 &5.38 &.724 &{}
        &.059 &1.98 &.201 \\
    
    \cmidrule{2-27}
    \multicolumn{2}{c}{Avg @A} &{}
        &.565 &4.24 &.575 &{}
        &1.07 &5.87 &- &{}
        &.495 &3.51 &.804 &{}
        &.145 &2.23 &.332 &{}
        
        &.668 &6.04 &.686 &{}
        &.064 &2.13 &.165 \\
    
    \midrule 

    \multicolumn{2}{c}{Avg} &{}
        &.516 &3.50 &.483 &{}
        &.806 &5.07 &- &{}
        &.409 &2.87 &.644 &{}
        &.143 &2.06 &.283 &{}
        &.491 &4.30 &.501 &{}
        &.060 &1.81 &.138
        \\
    \midrule \midrule
    \end{tabular}
    \vspace{-18pt}
\end{table*}
\endgroup

\section{Post-Flight Evaluation via ADR-FGO}
\label{sec:adr_fgo}
ADR-FGO is an offline, global full-batch factor-graph optimization framework~\cite{dellaert2017factor_graph} designed to provide high-fidelity reference trajectories for evaluation and analysis of the real-time ADR-VINS estimates. 
By minimizing global reprojection errors across the entire flight, ADR-FGO enforces trajectory-wide consistency and improves state estimates through batch smoothing. 
This results in a refined reference trajectory for benchmarking performance in uninstrumented environments. 
The proposed graph structure of ADR-FGO is illustrated in Fig. \ref{fig:adr_fgo_graph}.

\subsection{Factor Formulations}
The optimization problem is composed of three primary factor types that constrain the state vector at keyframe $k$, defined as $\mathbf{x}_k = [\mathbf{p}_k, \mathbf{v}_k, \mathbf{q}_k, \mathbf{b}_{a,k}, \mathbf{b}_{\omega,k
}]^\top$.

\subsubsection{IMU Preintegration Factors}
We utilize IMU preintegration on the $SO(3)$ manifold~\cite{forster2016manifold_imu_preintegration} to capture the relative motion ($\Delta \mathbf{p}_{k,k+1}, \Delta \mathbf{v}_{k,k+1}, \Delta \mathbf{q}_{k,k+1}$) between keyframes without repeated integration. The residual $\mathbf{r}_{IMU}$ is defined as:
\begin{equation}
    \mathbf{r}_{IMU,k} = \begin{bmatrix} \mathbf{R}_{k}^\top (\mathbf{p}_{k+1} - \mathbf{p}_{k} - \mathbf{v}_{k} \Delta t - \frac{1}{2}\mathbf{g} \Delta t^2) - \Delta \mathbf{p}_{k,k+1} \\ \mathbf{R}_{k}^\top (\mathbf{v}_{k+1} - \mathbf{v}_{k} - \mathbf{g} \Delta t) - \Delta \mathbf{v}_{k,k+1} \\ \log(\Delta \mathbf{q}_{k,k+1}^\top \otimes (\mathbf{q}_{k}^\top \otimes \mathbf{q}_{k+1})) \end{bmatrix}
\end{equation}

\subsubsection{Gate Corner Visual Factors}
These factors minimize the reprojection error of measurement $i$ relative to the known, fixed gate map $G$:
\begin{equation}
    \mathbf{r}_{corner,k,i} = \mathbf{\bar{u}}_{k,i} - h(\mathbf{x}_k, \mathbf{p}_{G,i}^\mathcal{W})
\end{equation}

\subsubsection{ADR-VINS Prior Factors}
We anchor the global optimization to the real-time ADR-VINS results using soft priors to improve convergence:
\begin{equation}
    \mathbf{r}_{prior,k} = \mathbf{x}_k \ominus \bar{\mathbf{x}}_k^{VINS}
\end{equation}

\subsubsection{Extrinsics Prior Factor}
This factor refine the camera-IMU transformation $T_{bc} = (R_{bc}, p_{bc})$ to account for mechanical shifts caused by impacts. 
It also allows for the refinement of nominal extrinsics when exact extrinsics are unavailable, as in the TII-RATM dataset. 
Specifically, we refine the rotational extrinsics $R_{bc}$, the residual $\mathbf{r}_{ext}$ is defined as:
\begin{equation}
  \mathbf{r}_{ext} = \text{Log}(\bar{R}_{bc}^\top R_{bc})  
\end{equation}
where $\text{Log}(\cdot)$ maps $SO(3)$ to $\mathfrak{so}(3)$.

\subsection{Keyframe Selection}
ADR-FGO generates a new keyframe at time $k$ if there is available visual measurements or a predefined time threshold $\tau_t$ has reached since the last keyframe, defined as:
\begin{equation}
  (\mathcal{Z}_k \neq \emptyset) \lor (t_{k} - t_{k-1} \geq \tau_t)  
\end{equation}
The second condition triggers "visual-less" keyframes.
It enables the optimizer to distribute accumulated drift and IMU biases across these keyframes, ensuring more accurate interpolation within segments lacking visual measurements.

\subsection{Factor Graph Optimization}
The final optimization minimizes a joint cost function over a set of keyframes $\mathcal{K}$. 
The joint cost function is defined as:
\begin{equation}
\begin{aligned}
    \min_{\mathcal{X}} & \left( \sum_{k \in \mathcal{K}} \| \mathbf{r}_{IMU,k} \|_{\Sigma_{IMU}}^2 + \sum_{k \in \mathcal{K}} \sum_{i \in \mathcal{I}_k}  \| \rho \left(\mathbf{r}_{corner,k,i}\right) \|_{\Sigma_{corner}}^2 \right. \\
    & \left. + \sum_{k \in \mathcal{K}} \| \mathbf{r}_{prior,k} \|_{\Sigma_{prior}}^2 + \| \mathbf{r}_{ext} \|_{\Sigma_{ext}}^2 \right)
\end{aligned}
\end{equation}
where $\Sigma_{IMU} \in \mathbb{R}^{9 \times 9}$, $\Sigma_{corner} \in \mathbb{R}^{2 \times 2}$, $\Sigma_{prior} \in \mathbb{R}^{6 \times 6}$, and $\Sigma_{ext} \in \mathbb{R}^{3 \times 3}$ are the weighting matrices which define the relative importance of each constraint in the graph.
$\rho(\cdot)$ is Huber function to handle detection noise.

\begin{figure*}
\begin{center}
    \includegraphics[width=1.0\textwidth]{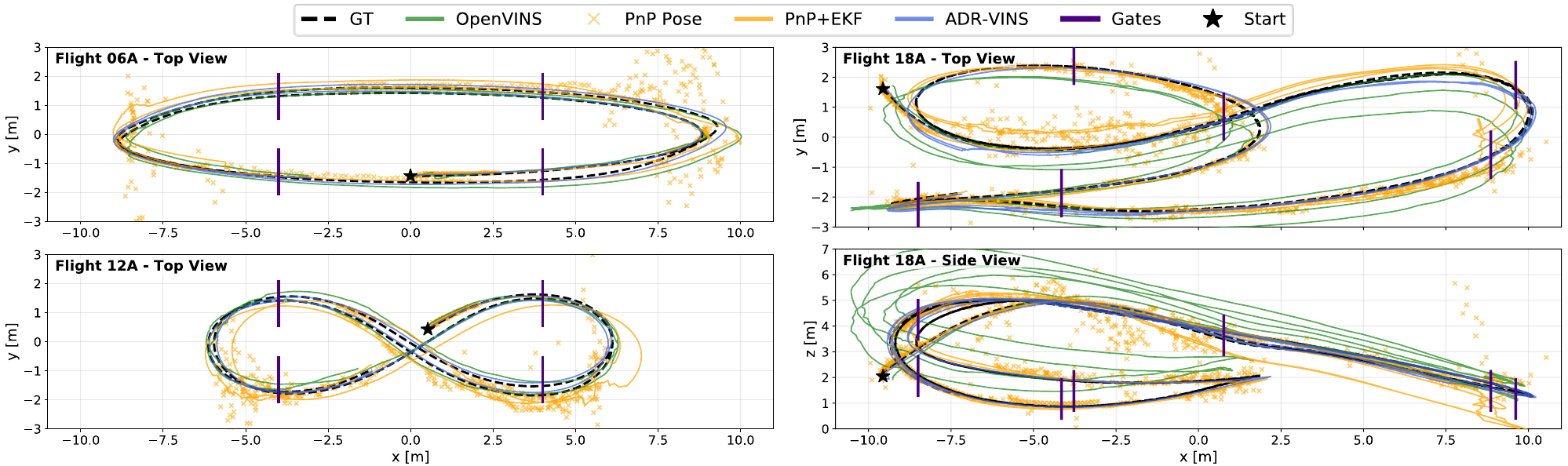}
    \vspace{-20pt}
    \caption{Qualitative comparison of ADR-VINS against baseline online methods across various tracks on the TII-RATM dataset autonomous sequences, including an ellipse (Flight 06A), a lemniscate (Flight 12A), and a complex 3D racing track (Flight 18A).}
    \label{fig:tii_results}
    \vspace{-18pt}
\end{center}
\end{figure*}

\section{Experiments}
We evaluate the proposed framework through benchmarks on the TII-RATM dataset and real-world closed-loop flights during the A2RL Drone Championship Season 2. The following sections detail our implementation, dataset results, and competition deployment.

\subsection{Implementation Details}
In ADR-VINS, the detection is handled by RTMO-nano~\cite{lu2024rtmo} optimized via TensorRT with FP16 precision with the ESKF uses Eigen3 library. 
The ADR-FGO utilizes SymForce~\cite{Martiros-RSS-22-symforce} for jacobian computation and factor-graph optimization.

\subsection{Metrics}
We mainly report the root mean squared (RMS) of translation ($e_t$, [m]), rotation ($e_r$, [deg]), and velocity ($e_v$, [m/s]) errors.
We also report the mean of pixels for detection and reprojection errors.

\subsection{TII-RATM Dataset Experiments}
\label{sec:tii_ratm_experiment}
\textbf{Description.} TII-RATM dataset~\cite{bosello2024tii_ratm} provides 500 Hz IMU and 120 Hz monocular RGB camera data from a drone as well as its 275 Hz ground-truth from a MoCap system. The indoor arena measuring $25 \times 9.7 \times 7$ m, where the drone was flown human-piloted and autonomously at $21.8$ m/s top speeds. 
Following ~\cite{bahnam2025iros}, we use sequences 05P, 06P, 11P, 12P, 05A, 06A, 11A, 12A, 17A, and 18A for our reports. 
Other sequences are used for training the detection model and tuning the baselines and our methods.
We limit the human-piloted flights to the first 3-laps.

\begin{figure}[t!]
\begin{center}
\includegraphics[width=1.0\columnwidth]{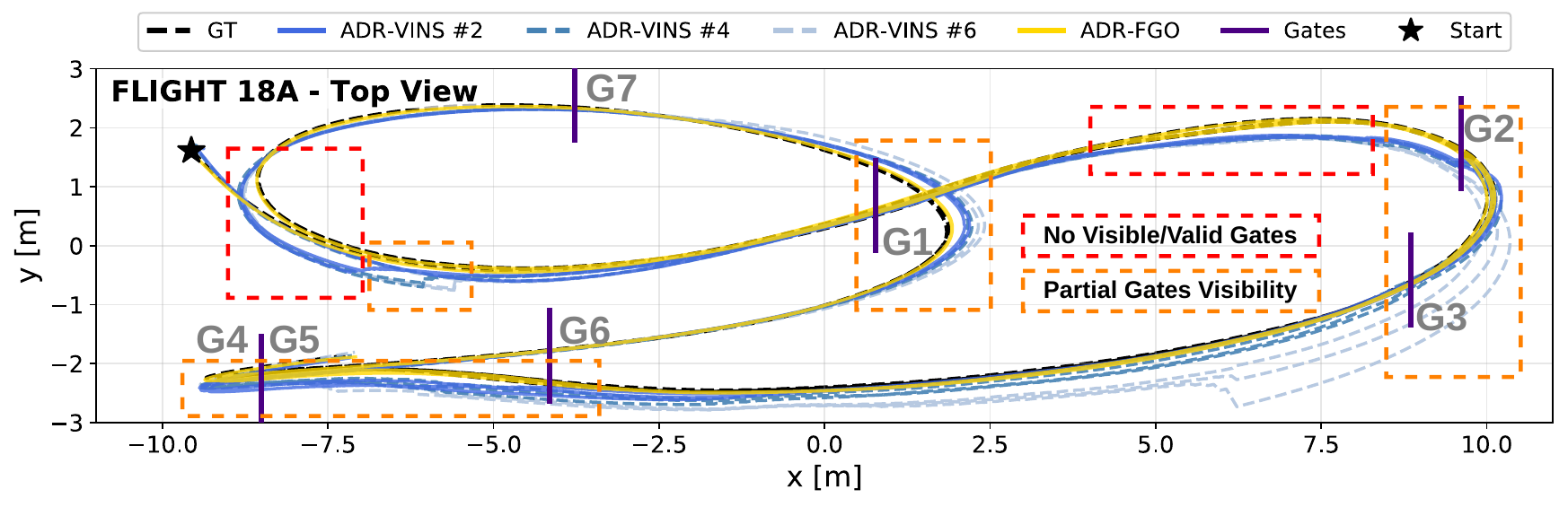}
\vspace{-20pt}
\caption{Comparison of ADR-VINS with $\#$2, $\#$4, and $\#$6 minimum corners visibility requirements for Flight 18A. Requiring only two corners enables earlier and more frequent updates, significantly reducing drift. ADR-FGO further minimizes this error by globally refining ADR-VINS, even in sections where gates are not visible. G1-G7 is the gate ordering.}
\label{fig:corners_and_adr_fgo}
\end{center}
\vspace{-16pt}
\end{figure}

\textbf{Baselines.} We benchmark ADR-VINS against OpenVINS~\cite{geneva2020openvins} and our own implementation of PnP+EKF~\cite{bahnam2026monorace} (including its RANSAC-based filtering) using the detection module explained in Sec. \ref{sec:gate_corner_detection}-\ref{sec:gate_association}.
For offline method, we benchmark ADR-FGO against MAPLAB\cite{cramariuc2022maplab}, a framework that performs full-batch optimization and global loop-closure using OpenVINS for its initial state guess.

\textbf{Initialization.} We initialize all online methods using ground-truth states because poor starting positions in TII-RATM dataset hindered initial gate detection for PnP+EKF and ADR-VINS, while OpenVINS failed to initialize on autonomous sequences. 
In real-world racing, these are resolved via hand-held motions to initialize OpenVINS pre-race (similar to T265-VIO~\cite{bosello2025on_your_own}).
For ADR-VINS, it can be statically placed on a podium with full visibility of the first gate for initial gate-based correction.

\begin{table}[t!]
    \centering
    \vspace{6pt}
    \caption{ADR-FGO Ablation Study.}
    \label{tab:adr_fgo_ablation}
    \setlength{\tabcolsep}{3.5pt}
    \vspace{-10pt}
    \begin{tabularx}{\columnwidth}{ccccccccccccc}
    \midrule \midrule
    &{} &\multicolumn{3}{c}{w/o vl KF} & &\multicolumn{3}{c}{w/o ext factor} & &\multicolumn{3}{c}{Full} \\
    Seq. &{} &$e_t$ &$e_r$ &$e_v$ & &$e_t$ &$e_r$ &$e_v$ & &$e_t$ &$e_r$ &$e_v$ \\
    \midrule
    \multirowcell{1}{Avg @P} &{} &.056 &1.51 &.111 & &.067 &1.88 &.115 & &.054 &1.32 &.098\\
    \multirowcell{1}{Avg @A} &{} &.064 &2.37 &.177 & &.070 &2.15 &.179 & &.064 &2.13 &.165\\
    \midrule \midrule
    \end{tabularx}
    \begin{itemize}
        \item w/o vl KF: without visual-less keyframes. 
        \item w/o ext factor: without extrinsics factor.
    \end{itemize}
    \vspace{-16pt}
\end{table}

\textbf{ADR-VINS Results.}
As shown in Table \ref{tab:tii_results}, ADR-VINS achieved an average translational RMSE of $0.141$ m and $0.145$ m in piloted and autonomous flights, respectively.
It significantly outperforms OpenVINS and PnP+EKF, reducing translational, rotational, and velocity errors by $65\%$--$72\%$, $28\%$--$41\%$, and $41\%$--$56\%$, respectively. 
Fig. \ref{fig:tii_results} shows ADR-VINS tracks ground truth closely even during aggressive 3D maneuvers (e.g., Flight 18A) where baselines drift substantially. 
Note that PnP solvers are prone to noise and initial guess, leading to high raw PnP-Pose errors.     
While the PnP+EKF fusion mitigates these errors by $49\%$, it remains uncompetitive with ADR-VINS.

\textbf{Partial Gate Updates.}
Fig. \ref{fig:corners_and_adr_fgo} evaluates corner visibility constraints during Flight 18A. 
ADR-VINS achieves highest accuracy ($0.208$ m) when requiring just two visible corners per update. 
Increasing this minimum to four (the PnP standard) or six corners (as used in \cite{bahnam2026monorace}) degrades the $e_t$ to $0.228$ m and $0.387$ m, respectively.
This shows that partial gate updates enable earlier, more frequent corrections, maintaining robust tracking even when aggressive maneuvers frequently make the gates out of sight.

\begin{figure}[t!]
\centering
\includegraphics[width=1.0\columnwidth]{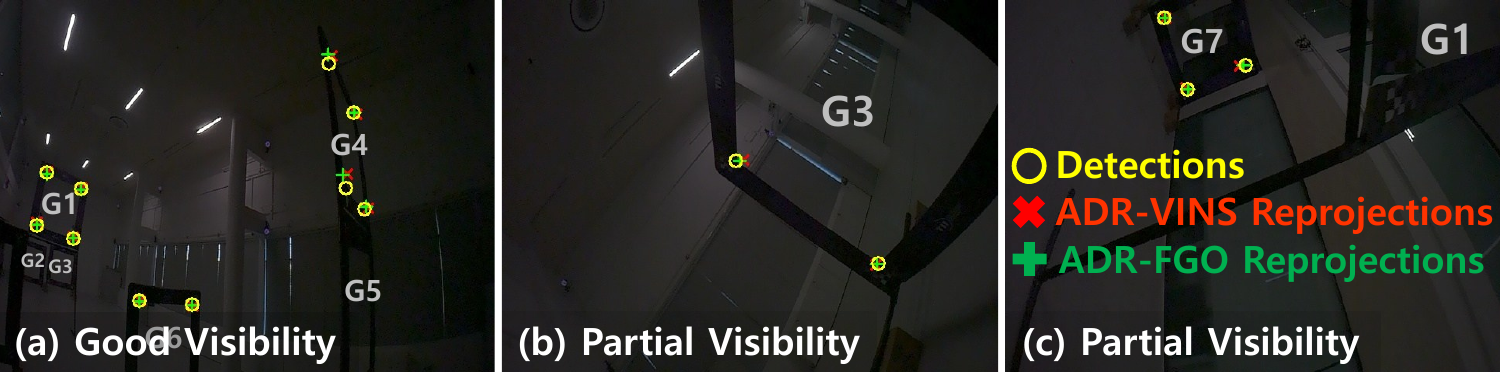}

\vspace{4pt}

\includegraphics[width=1.0\columnwidth]{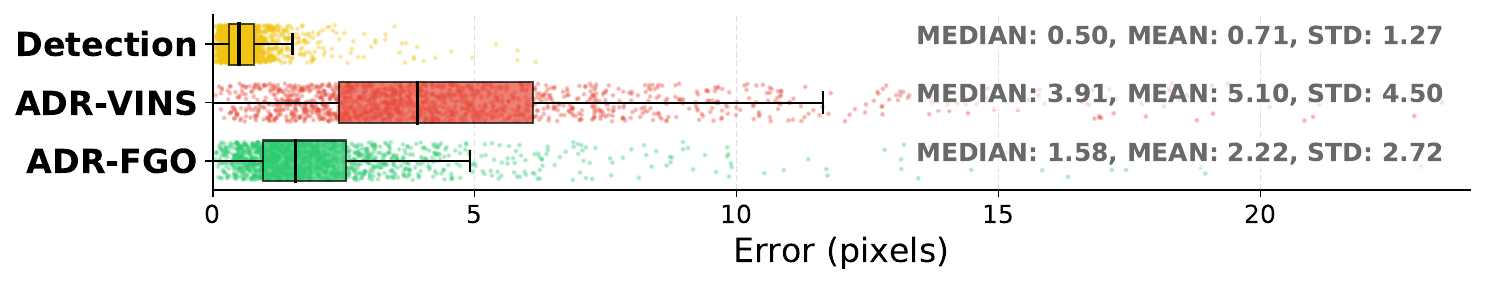}

\vspace{-12pt}
\caption{Top: Visualization of reprojection errors for ADR-VINS and ADR-FGO compared to raw detections on flight 18A. Bottom: Statistics of the detection and reprojection errors on TII-RATM dataset.}
\label{fig:detections_reprojections_tii}
\vspace{-18pt}
\end{figure}

\textbf{ADR-FGO Results.}
As shown in Table \ref{tab:tii_results}, ADR-FGO achieves an average translational error of $0.060$ m, significantly more accurate compared to MAPLAB with translation error of $0.491$ m, $88\%$ relative improvement. 
This is largely due to MAPLAB's failure in establishing loop-closures during high-speed flights.
The $58\%$ translational accuracy improvement against ADR-VINS makes ADR-FGO suitable as the reference trajectory for evaluating and tuning ADR-VINS.
As shown in Fig. \ref{fig:corners_and_adr_fgo}, ADR-FGO produces smooth reference paths, recovering the trajectory even in sections without corner detections (red area). 
An ablation study in Table III further highlights that utilizing visual-less keyframes and refining camera-IMU extrinsics are beneficial.

\textbf{Reprojection Errors.}
Image-space reprojection errors can serve as a proxy for evaluating state estimation performance. 
As shown in Fig. \ref{fig:detections_reprojections_tii}, the detection module has an error of $0.71$ pixels, ADR-VINS maintains a $5.1$ pixels reprojection errors in real-time, effectively bounding the drift. 
ADR-FGO further refines these estimates, halving the reprojection errors to $2.22$ pixels. 
ADR-FGO's refined reprojections align closely with the detections, maintaining high fidelity even during aggressive maneuvers and partial gate visibility.

\begin{figure}[ht!]
\begin{center}
\centering
\includegraphics[width=0.8\columnwidth]{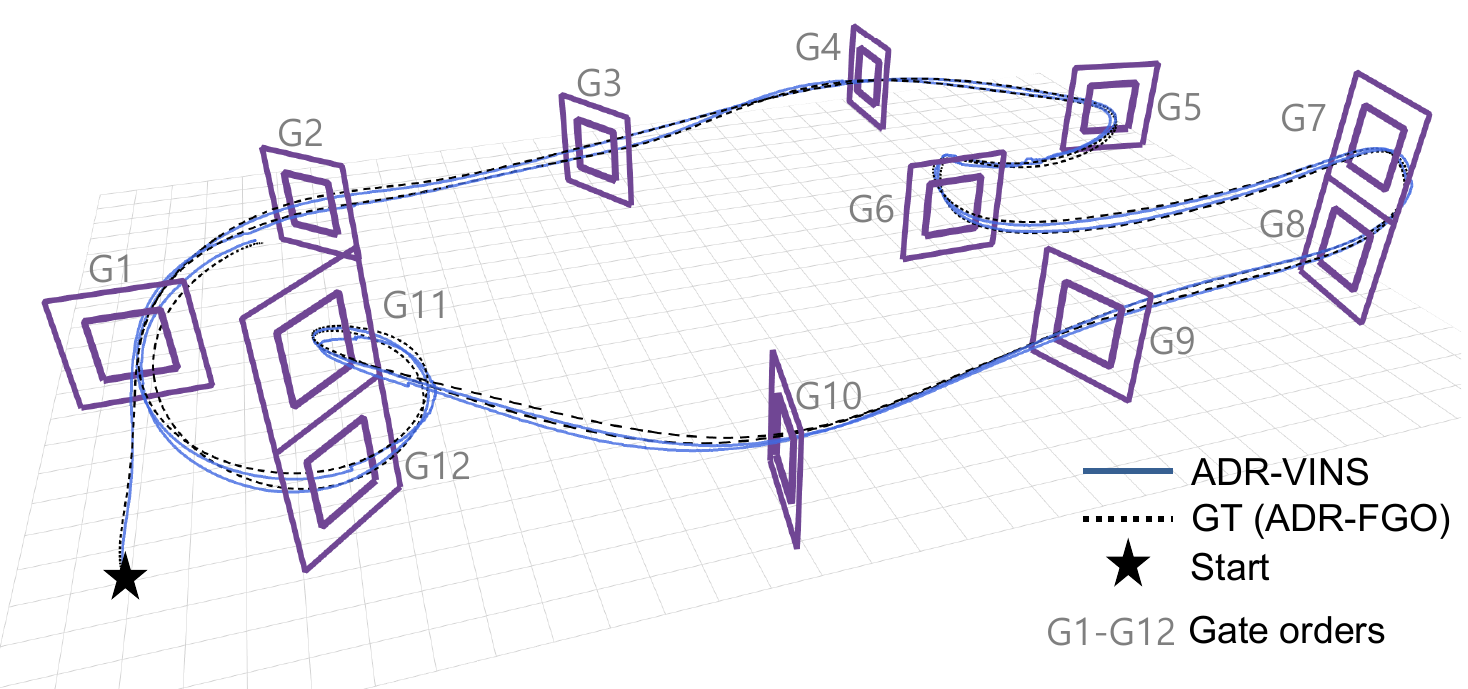}

\vspace{0pt}

\includegraphics[width=1.0\columnwidth]{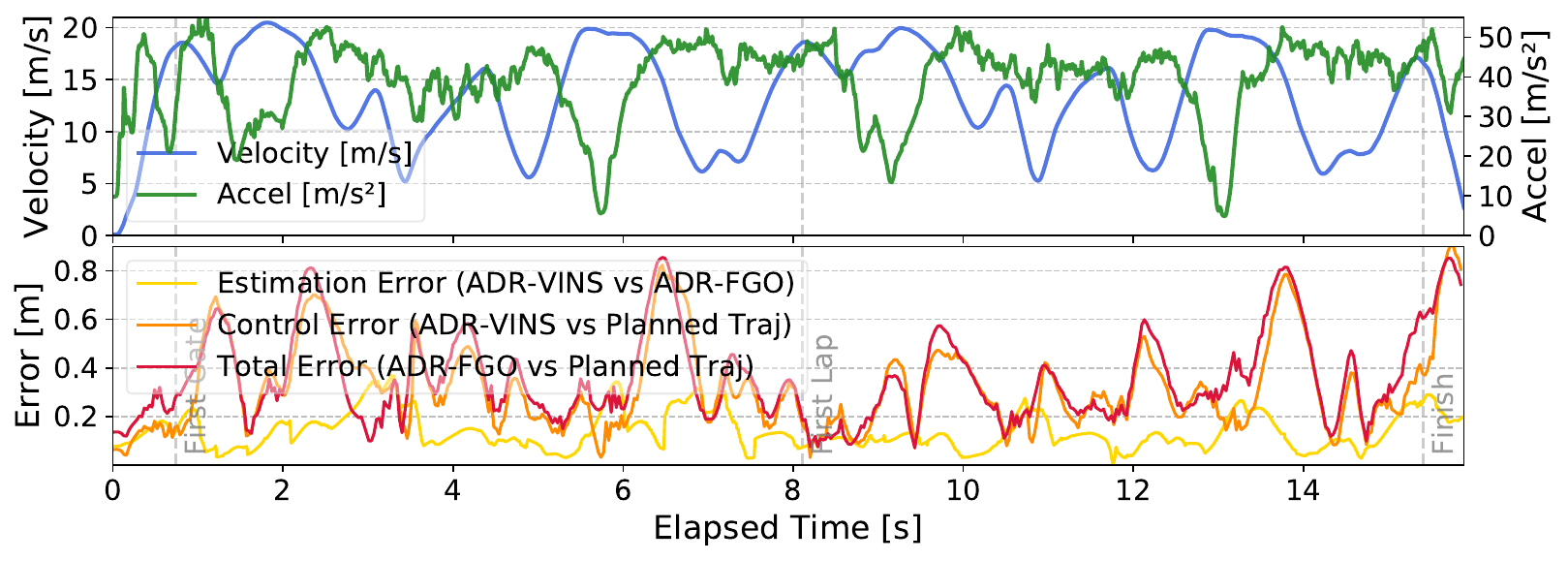}

\vspace{-12pt}
\caption{Top: 3D visualization of ADR-VINS against GT (ADR-FGO) of our fastest flight in A2RL S2 track. Middle: Velocity and acceleration profile. Bottom: Error analysis.}
\label{fig:a2rl_result}
\end{center}
\vspace{-12pt}
\end{figure}

\subsection{A2RL Drone Championship Season 2 Deployment}
\textbf{Description.}
The system was deployed in the A2RL Drone Championship Season 2 in a competitive, uninstrumented racing track. 
The track ($35 \times 20 \times 7$ m) consisted of 12 gates per-lap, including difficult \textit{Split-S} and \textit{ladder-down} maneuvers.
All development, tuning, and experiments prior to the competition were conducted without the use of any on-premise MoCap systems, relying solely on onboard sensors and the known gate map.
Furthermore, as MoCap system is not available in the competition site, we utilize ADR-FGO to generate a reference trajectory for evaluation and analysis.

\textbf{Platform.}
Our real-world deployment utilizes a $30 \times 28 \times 10$ cm, $960$ g racing drone platform standardized for A2RL S2, as shown in Fig. \ref{fig:fig1}. 
Perception is provided by an Arducam IMX219 rolling shutter camera ($820 \times 616$ at $85$ Hz) pitched at $\sim45^\circ$ with $155^\circ\times115^\circ$ FOV. 
The platform features an Jetson Orin NX computer interfaced with a Foxeer H7 flight controller (FC) via UART/MSP, receiving $500$ Hz IMU data and sending $100$ Hz control CTBR commands.
We employ an MPC controller~\cite{Falanga2018pampc} to track the planned time-optimal trajectory from TOGT~\cite{qin2024togt}.

\begin{table}[t!]
    \centering
    \caption{Results on A2RL Season 2 Track.}
    \label{tab:result_on_a2rl_track}
    \setlength{\tabcolsep}{5.7pt}
    \vspace{-8pt}
    \begin{tabularx}{\columnwidth}{cccccccccccc}
    \midrule \midrule
    \multicolumn{3}{c}{PnP-Pose} & &\multicolumn{3}{c}{PnP+EKF} & &\multicolumn{3}{c}{ADR-VINS} \\
    $e_t$ &$e_r$ &$e_v$ & &$e_t$ &$e_r$ &$e_v$ & &$e_t$ &$e_r$ &$e_v$ \\
    \midrule
    .797 &5.97 &- &{} &.367 &2.31 &.717 &{} &.152 &1.25 &.413\\
    \midrule \midrule
    \end{tabularx}
    \vspace{-18pt}
\end{table}

\textbf{Results.}
Fig. \ref{fig:a2rl_result} shows the visualization of our fastest attempt at A2RL Season 2. 
ADR-VINS remained stable even during aggressive maneuvers like \textit{Split-S} and \textit{Ladder-down}, with top speed reaching $20.9 \text{ m/s}$ and acceleration over $5 g$. 
Based on the refined trajectory (ADR-FGO), ADR-VINS has translational RMSE of $0.152$ m.
While the control tracking error is $0.370$ m and the total error (ADR-FGO vs planned trajectory) is $0.402$ m.
Note that the main contributor to the total error is the control tracking error.
Furthermore, as shown in Table \ref{tab:result_on_a2rl_track}, ADR-VINS significantly outperforms the baselines, reducing the translational error by $58\%$ and the rotational error by $46\%$ compared to the PnP+EKF implementation.

\begin{figure}[t!]
\centering
\includegraphics[width=1.0\columnwidth]{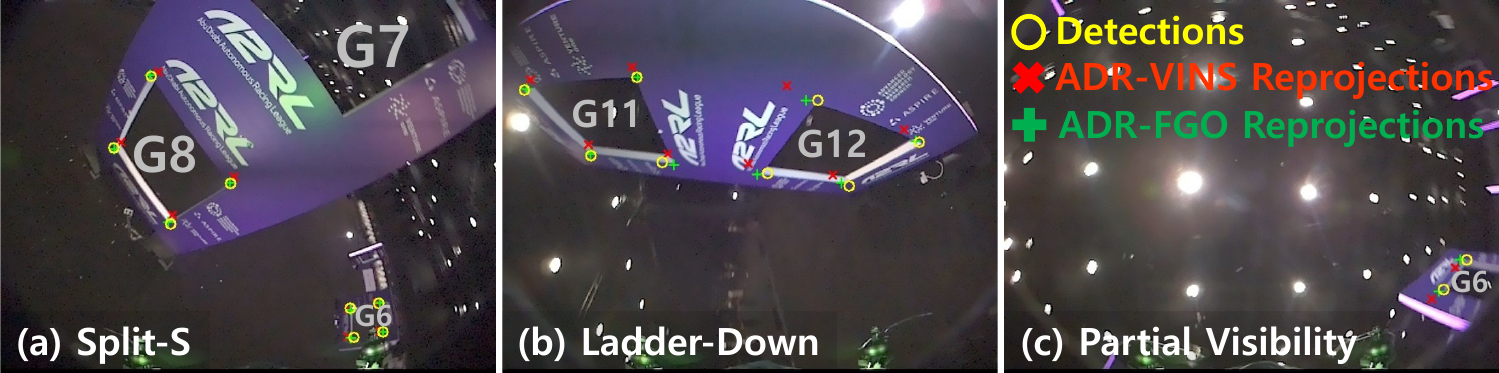}
\vspace{-12pt}
\includegraphics[width=1.0\columnwidth]{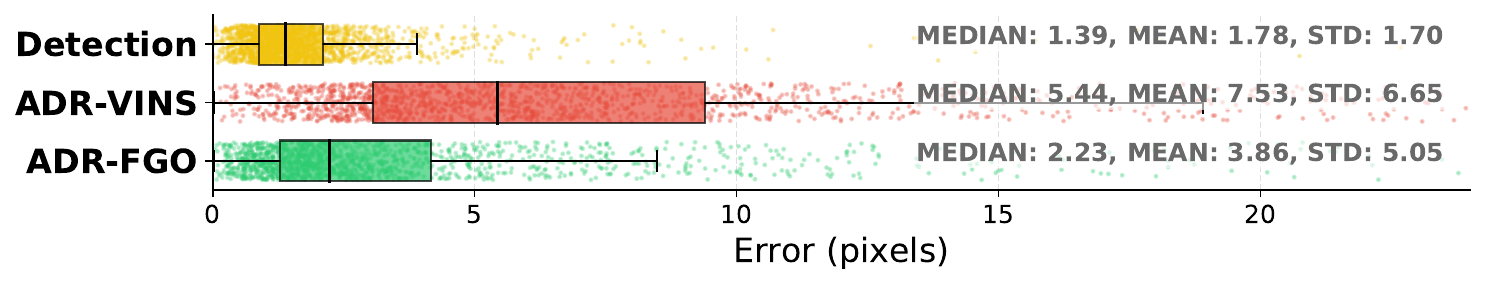}
\vspace{-12pt}
\caption{Top: Visualization of the corner detections and the reprojections of ADR-VINS and ADR-FGO estimates. Bottom: Statistics of the detection and reprojection errors on our fastest flight in A2RL S2.}
\label{fig:detections_reprojections_a2rl}
\vspace{-6pt}
\end{figure}

\textbf{Reprojection Errors.}
Fig. \ref{fig:detections_reprojections_a2rl} compares how well the estimated drone position aligns with the actual gate corners seen by the camera. In this environment, ADR-VINS achieved a reprojection error of $7.53$ pixels and ADR-FGO refines it to $3.86$ pixels. 
This higher error compared to Sec. \ref{sec:tii_ratm_experiment} is due to an increase in detection noise, camera noise, and limited training data, yet it was able to robustly provide accurate state-estimation. 

\begin{figure}[t!]
\centering
\includegraphics[width=0.8\columnwidth]{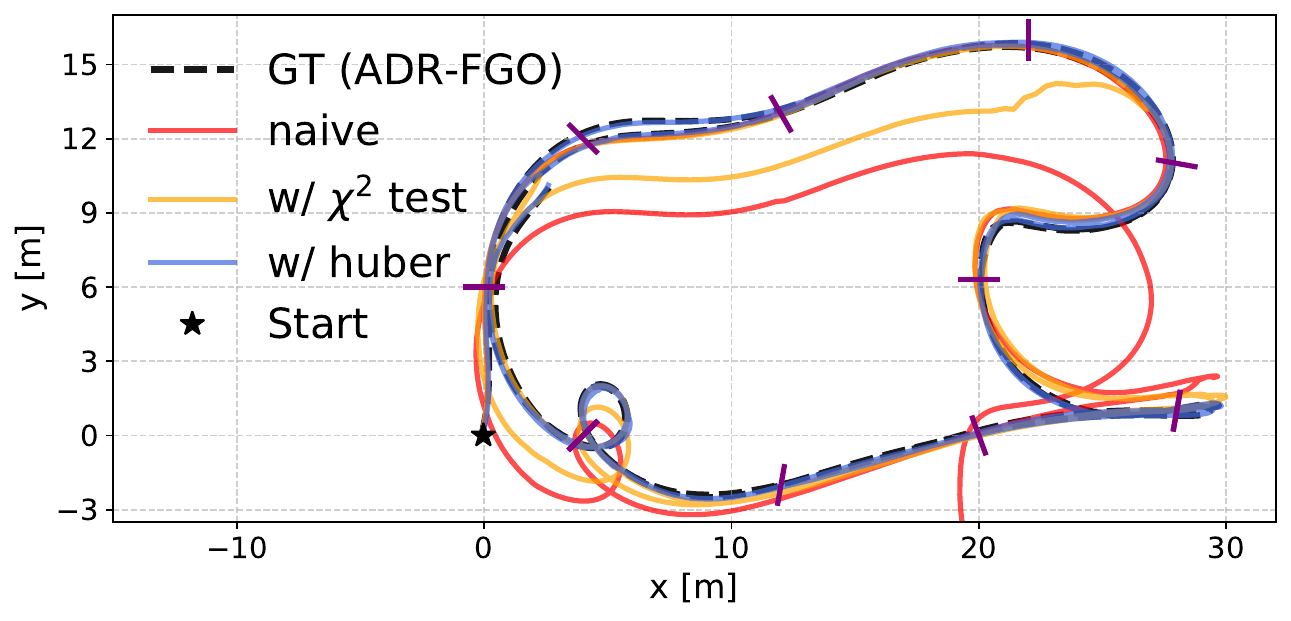}
\vspace{-12pt}
\caption{Robustness ablation study visualization. Performance comparison of the proposed Huber-based reweighting, a $\chi^{2}$ (chi-squared) rejection test, and naive visual measurements fusion. The Huber-based method maintains stability against noisy detections where naive and $\chi^{2}$ test variants exhibit significant drift or failure.}
\label{fig:robustness_ablation}
\vspace{-12pt}
\end{figure}

\textbf{Robustness.}
The benefit of our robust reweighting method is highlighted in Fig. \ref{fig:robustness_ablation}.
We ablate the way we fuse the noisy detections in the filter.
A "naive" version of ADR-VINS that treats all measurements equally quickly failed and drifted away outside the track. 
A $\chi^2$ rejection is slightly more stable but resulted in sections with significant drift. 
The Huber-based approach allowed the drone to ignore noisy data points while still using helpful information, resulting in a flight path that closely matched the high-accuracy reference.

\begin{table}[t!]
    \centering
    \vspace{2pt}
    \caption{Timing of ADR-VINS.}
    \label{tab:runtime_adr_vins}
    \setlength{\tabcolsep}{4.8pt}
    \vspace{-8pt}
    \begin{tabular}{rcccccc}
    \midrule \midrule
    &{} &\multicolumn{2}{c}{Visual Measurements} &{} &\multicolumn{2}{c}{ESKF} \\
    \cmidrule{3-4} \cmidrule{6-7}
    component: &{} &detection &reorder.+assoc. &{} &propagation &update \\
    \midrule
    time [ms]: &{} &8.165 &0.192 &{} &0.011 &0.242 \\
    \midrule \midrule
    \end{tabular}
    \label{tab:runtime}
    \vspace{-18pt}
\end{table}

\textbf{Runtime.}
As detailed in Table \ref{tab:runtime_adr_vins}, ADR-VINS operates under $9$ ms per frame, meeting the $85$ Hz ($11.7$ ms) camera's real-time constraints. 
The RTMO-nano detection module dominates latency, while all remaining processes execute in sub-millisecond times. 
For comparison, in PnP+EKF, the RANSAC+PnP computation adds a net $0.4$ ms latency, about $200\%$ of the reordering+association time. 
Future improvements could reduce detection latency via smaller model or by upgrading to JetPack 6.2 to leverage an increased computational power budget. 
Finally, the offline ADR-FGO process takes about 15–60 seconds, depending on trajectory length and optimization convergence.

\section{Conclusion and Discussion}
\textbf{Conclusion.}
We presented ADR-VINS, a tightly-coupled ESKF framework for high-speed drone racing, combined with ADR-FGO for offline evaluation. 
The proposed system demonstrated high precision on the TII-RATM dataset and proven reliability during the A2RL Drone Championship Season 2 at speeds up to $20.9~m/s$.
We also showed the utilization of ADR-FGO in evaluating the estimation, control, and system performance in uninstrumented environment.

\textbf{ADR-VINS Discussion.}
Despite its successful deployment, ADR-VINS is currently constrained by the label quality and limited training data for its real-world deployment. 
Like prior methods, ADR-VINS assumes a known gate map, which is typically available prior to the competitions.
Furthermore, it utilizes only inner corners instead of both inner and outer corners~\cite{bahnam2026monorace} for improved redundancy and accuracy. 
Nevertheless, the proposed filter maintains stable estimation despite these shortcomings.
Synthetic dataset generation~\cite{jacinto2024pegasus_sim} could improve the detection performance.
Furthermore, extended periods without visual measurements, inherent IMU drift could be a challenge; potentially, learning-based motion predictions~\cite{cioffi2023imo,bahnam2025iros} could provide a more accurate state propagation.

\textbf{ADR-FGO Discussion.}
While ADR-FGO provides high-fidelity accuracy for post-flight analysis and fine-tuning, its improvement in rotational accuracy over ADR-VINS is limited by the sparse gate features available. 
In contrast, frameworks like MAPLAB~\cite{cramariuc2022maplab} utilize many more visual features for improved orientation estimates but it is unreliable at very high speed flights. 
Beyond trajectory improvement, ADR-FGO refines camera-IMU extrinsics, which could be used to improve extrinsics post-crash, as exemplified in ~\cite{bahnam2026monorace}. 
In the future, the refined trajectories can serve as supervision for training data-driven motion models~\cite{cioffi2023imo, bahnam2025iros} to further improve ADR-VINS state propagation when corner detections are unavailable.

\textbf{System Performance.}
Lastly, system-wide performance is currently limited by the tracking performance of the MPC and modeling inaccuracies, as tracking error dominates the total system error. 
Recent study~\cite{song2023oc_vs_rl} and the success of the A2RL S1 and S2 winners suggest that compared to MPC, RL-based control offers superior robustness and accuracy needed for high-speed racing at the performance's edge.



\section{Acknowledgement}
We thank June-Kyoo Park for his contributions as a member of Team KAIST during A2RL Season 2, and we are grateful to A2RL, ASPIRE, and everyone involved for their excellent support throughout the championship.
We acknowledge the use of Gemini 3 for coding assistance and manuscript refinement, specifically to improve grammar and flow.
\begingroup
\bibliographystyle{ieeetr}
\bibliography{root}
\endgroup

\clearpage

\begin{strip}
\centering
\includegraphics[width=1.0\textwidth]{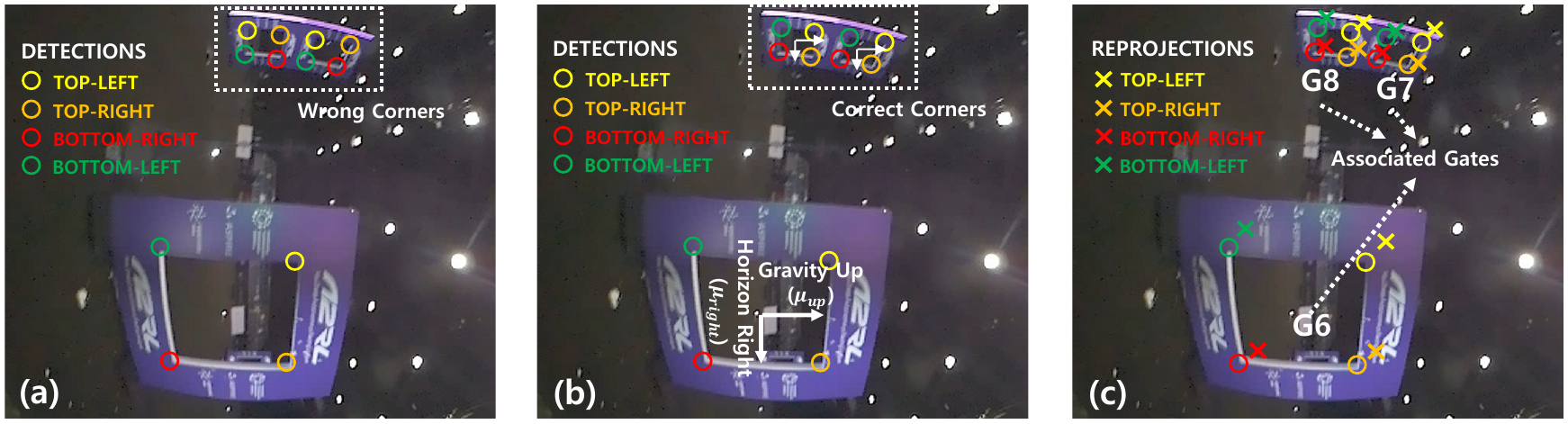}
\vspace{-22pt}
\captionof{figure}{Visualization of the visual measurement module detailed in Section \ref{sec:gate_corner_detection} - \ref{sec:gate_association}. (a) The detection module detect the each gate with its corresponding corners (TOP-LEFT, TOP-RIGHT, BOTTOM-RIGHT, and BOTTOM-LEFT). Due to visual ambiguities at high roll angle and far-away gates, the corner label predictions can be wrong. (b) Based on the ADR-VINS state, we reproject the gravity vector up and right for each gate centroid to the image plane and reorder the corners based on Sec. \ref{sec:corner_reordering} to obtain the correctly assigned gate corners. (c) We reproject the known gate map onto the image using the ADR-VINS state, which are then associated to the detection based on Sec.\ref{sec:gate_association}.}
\label{fig:visual_measurement_vis}
\vspace{-12pt}
\end{strip}

\appendix
\subsection{Visual Measurement Visualization}
The visualization is shown in Fig. \ref{fig:visual_measurement_vis}.


\begin{figure}[h!]
\centering
\includegraphics[width=1.0\columnwidth]{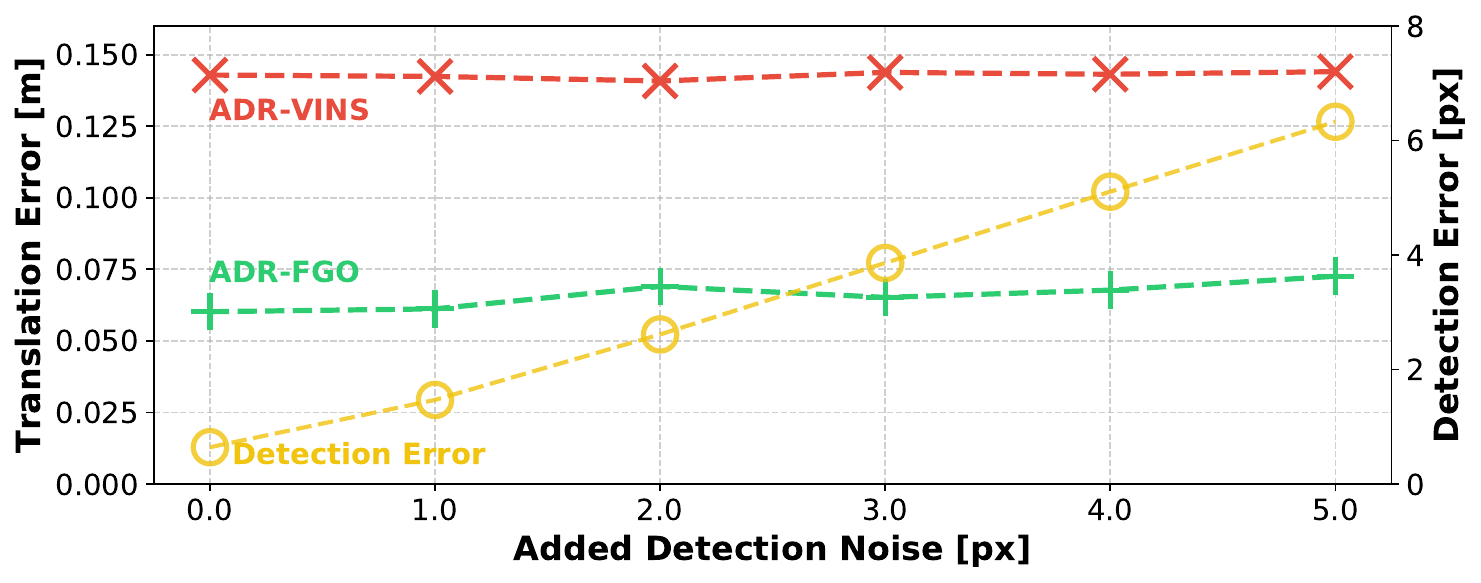}
\vspace{-22pt}
\caption{Robustness against added detection noise in TII-RATM Dataset.}
\label{fig:robustness_to_detection}
\vspace{-12pt}
\end{figure}

\subsection{Robustness to Detection Noise}

\textbf{Description.}
The robustness of the proposed frameworks against perceptual inaccuracies is evaluated on the TII-RATM dataset. 
As a proxy for the perceptual degradation seen in deployment, zero-mean Gaussian noise ($0.0$ to $5.0$ pixels) was added to the corner detections. 
The resulting performance, shown in Fig. \ref{fig:robustness_to_detection}, compares the translation error of the online ADR-VINS and the offline ADR-FGO against increasing detection error. 
In practice, however, this noise often departs from the zero-mean Gaussian assumption, such as image noise, motion blur, and rolling shutter can add biases. 

\textbf{ADR-VINS Results.}
The results show that ADR-VINS maintains a stable translation error of approximately $0.14\text{ m}$ across the testing range. 
This stability stems from the integration of Huber-based robust reweighting within the ESKF which downweights noisy innovation terms while preserving useful measurement data. 

\textbf{ADR-FGO Results.}
Furthermore, ADR-FGO consistently provides a "silver-standard" reference with a translation error between $0.06\text{ m}$ and $0.073\text{ m}$. 
While ADR-FGO exhibits slight sensitivity to extreme noise, it remains significantly more accurate than ADR-VINS. 
These findings validate the system's reliability to be utilized for evaluation and analysis purposes in uninstrumented environments where accurate detection can be hard to achieve.




\end{document}